\documentstyle[eaclap]{article}

\newcommand {\bc}{\begin{center}}
\newcommand {\ec}{ \end{center}}
\newcommand {\be} {\begin{enumerate}}
\newcommand {\ee} {\end{enumerate}}
\newcommand {\bex} {\begin{example}}
\newcommand {\eex} {\end{example}}
\newcommand {\bsex} {\begin{alphex}}
\newcommand {\esex} {\end{alphex}}
\newcommand {\attr}[1]{{ \sc{#1}}}

\newcommand{\logic}{{${\cal L}$}}
\newcommand{\mo}{{\bf M}}
\newcommand{\zi}{zoomin}
\newcommand{\ifff}{ \ \mbox{{\it iff }} \ }
\newcommand{\ftr}[1]{{\langle #1 \rangle}}
\newcommand{\ua}{\ftr{up}}
\newcommand{\upa}{\ua}

\newcommand{\da}{\ftr{down}}

\newcommand{\peq}{\approx}
\newcommand{\txt}[1]{{\mbox{\it #1}}}

\def\evnup{\@ifnextchar[{\@evnup}{\@evnup[0pt]}}
\def\@evnup[#1]#2{\setbox1=\hbox{#2}%
\dimen1=\ht1 \advance\dimen1 by -.5\baselineskip%
\advance\dimen1 by -#1%
\leavevmode\lower\dimen1\box1}

\title{\vspace{-0.5cm}
A Specification Language for \\
Lexical Functional Grammars}

\author{
Patrick Blackburn and Claire Gardent \\
Computerlinguistik \\
Universit\"{a}t des Saarlandes \\
Postfach 1150, D-66041 Saarbr\"{u}cken  \\
Germany\\
{\tt \{patrick,claire\}@coli.uni-sb.de}
}

\begin{document}

\bibliographystyle{acl}

\maketitle
\vspace{-0.5cm}
\begin{abstract}
This paper defines a language $\cal L$ for specifying LFG grammars.
This enables constraints on LFG's composite ontology (c-structures
synchronised with f-structures) to be stated directly; no appeal to
the LFG construction algorithm is needed.  We use $\cal L$ to specify
schemata annotated rules and the LFG uniqueness, completeness and
coherence principles.  Broader issues raised by this work
are noted and discussed.
\end{abstract}

\section{Introduction}
\label{0}

Unlike most linguistic theories, LFG (see Kaplan and Bresnan (1982)) treats
grammatical relations as first class citizens.  Accordingly, it casts its
linguistic analyses in terms of a composite ontology: two independent
domains --- a domain of constituency information (c-structure), and a
domain of grammatical function information (f-structure) --- linked
together in a mutually constraining manner.  As has been amply demonstrated
over the last fifteen years, this view permits perspicuous analyses of a
wide variety of linguistic data.

However standard formalisations of LFG do not capture its strikingly simple
underlying intuitions.  Instead, they make a detour via the LFG {\em
construction algorithm}, which explains how equational constraints linking
subtrees and feature structures are to be resolved.  The main point of the
present paper is to show that such detours are unnecessary.  We define a
specification language ${\cal L}$ in which (most of) the interactions
between c- and f-structure typical of LFG grammars can be stated directly.

The key idea underlying our approach is to think about LFG model
theoretically.  That is, our first task will be to give a precise --- and
{\em transparent} --- mathematical picture of the LFG ontology.  As has
already been noted, the basic entities underlying the LFG analyses are
composite structures consisting of a finite tree, a finite feature
structure, and a function that links the two.  Such structures can
straightforwardly be thought of as {\em models}, in the usual sense of
first order model theory (see Hodges (1993)).  Viewing the LFG ontology in
such terms does no violence to intuition: indeed, as we shall see, a more
direct mathematical embodiment of the LFG universe can hardly be imagined.

Once the ontological issues have been settled we turn to our ultimate goal:
providing a specification language for LFG grammars.  Actually, with the
ontological issues settled it is a relatively simple task to devise
suitable specification languages: we simply consider how LFG linguists talk
about such structures when they write grammars.  That is, we ask ourselves
what kind of constraints the linguist wishes to impose, and then devise a
language in which they can be stated.

Thus we shall proceed as follows.  After a brief introduction to LFG,%
\footnote{This paper is based upon the original
formulation of LFG, that of Kaplan and Bresnan (1982),
and  will not discuss such later innovations as functional
uncertainty.}
we isolate a class of models which obviously mirrors
the composite nature of the LFG ontology, and then turn to the task of
devising a language for talking about them.  We opt for a particularly
simple specification language: a propositional language enriched with
operators for talking about c- and f-structures, together with a path
equality construct for enforcing synchronisation between the two
domains.  We illustrate its use by showing how to
capture the effect of schemata annotated rules, and the LFG
uniqueness, completeness and coherence principles.

Before proceeding, a word of motivation is in order.  Firstly, we believe
that there are practical reasons for interest in grammatical specification
languages: formal specification seems important (perhaps essential) if
robust large scale grammars are to be defined and maintained.  Moreover,
the essentially model theoretic slant on specification we propose here
seems particularly well suited to this aim.  Models do not in any sense
``code'' the LFG ontology: they take it pretty much at face value.  In our
view this is crucial.  Formal approaches to grammatical theorising should
reflect linguistic intuitions as directly as possible, otherwise they run
the risk of being an obstacle, not an aid, to grammar development.

The approach also raises theoretical issues.  The model theoretic approach
to specification languages forces one to think about linguistic ontologies
in a systematic way, and to locate them in a well understood mathematical
space.  This has at least two advantages.  Firstly, it offers the prospect
of meaningful comparison of linguistic frameworks.  Secondly, it can
highlight anomalous aspects of a given system.  For example, as we shall
later see, there seems to be no reasonable way to deal with LFG's $=_c$
definitions using the simple models of the present paper.  There {\em is} a
plausible  model theoretic strategy strategy for extending our
account to cover $=_c$; but the nature of the required extension clearly
shows that $=_c$ is of a quite different character to the bulk of LFG.  We
discuss the matter in the paper's concluding section.

\section{Lexical Functional Grammar}

A lexical functional grammar consists of three main components: a set of
context free rules annotated with schemata, a set of well formedness
conditions on feature structures, and a lexicon.  The role of these
components is to assign two interrelated structures to any linguistic
entity licensed by the grammar: a tree (the {\it c-structure}) and a
feature structure (the {\it f-structure}).  Briefly, the context free
skeleton of the grammar rules describes the c-structure, the
well-formedness conditions restrict f-structure admissibility, and the
schemata synchronise the information contained in the c- and f-structures.

\begin{figure}[hbtp]
\bc
\begin{tabular}{lcccc}
(1) & S  &      $\longrightarrow$       & NP    & VP \\
                                & ($\uparrow$ \attr{subj}) = $\downarrow$  &
                                & $\uparrow$ = $\downarrow$ \\
    &&&&\\
(2) & NP &      $\longrightarrow$       & Det   & N \\
    &&&&\\
(3) & VP &      $\longrightarrow$       & V & \\
                                & $\uparrow = \downarrow$ &\\
&& \\
\end{tabular}

\begin{tabular}{ll}
$a$     & Det, ($\uparrow$ \attr{spec}) = a,
        ($\uparrow$ \attr{num}) = sing \\

$girl$  & N, ($\uparrow$ \attr{pred}) = girl,
        ($\uparrow$ \attr{num}) = sing \\

$walks$ & V, ($\uparrow$ \attr{pred}) = walk($\ftr{subj}$), \\
  & ($\uparrow$ \attr{tense}) = pst \\ & \\ \end{tabular} \ec
\caption{An LFG grammar fragment}
\label{f1} \end{figure}

To see how this works, let's run through a simple example.  Consider
the grammar given in Figure \ref{f1}.  Briefly, the up- and
down-arrows in the schemata can be read as follows: $\uparrow
\mbox{{\it Feature}}$ denotes the value of {\it Feature} in the
f-structure associated with the tree node immediately dominating the
current tree node, whereas $\downarrow Feature$ denotes the value of
$Feature$ in the f-structure associated with the current tree node.
For instance, in rule (1) the NP schema indicates that the f-structure
associated with the NP node is the value of the \attr{subj} feature in
the f-structure associated with the mother node.  As for the VP
schema, it requires that the f-structure associated with the mother
node is identical with the f-structure associated with the VP node.

Given the above lexical entries, it is possible to assign a correctly
interrelated c-structure and f-structure to the sentence {\it A girl
walks}.  Moreover, the resulting f-structure respects the LFG well
formedness conditions, namely the {\it uniqueness}, {\it completeness}
and {\it coherence} principles discussed in section 5.  Thus {\it A
girl walks} is accepted by this grammar.

\section{Modeling the LFG ontology}

The ontology underlying LFG is a composite one, consisting of trees,
feature structures and links between the two.  Our first task is to
mathematically model this ontology, and to do so as transparently as
possible.  That is, the mathematical entities we introduce should
clearly reflect the intuitions important to LFG theorising --- ``No
coding!'', should be our slogan.  In this section, we introduce such a
representation of LFG ontology.  In the following section, we shall
present a formal language \logic\ for talking about this representation;
that is, a language for specifying LFG grammars.

We work with the following objects.  A {\em model} \mo\ is a tripartite
structure $\langle {\cal T}, zoomin, {\cal F} \rangle$, where $\cal{T}$ is
our mathematical picture of c- structure, $\cal{F}$ our picture of
f-structure, and $zoomin$ our picture of the link between the two.  We now
define each of these components.  Our definitions are given with respect to
a {\em signature} of the form $\langle Cat, Atom, Feat
\rangle$, where {\it Cat}, {\it Atom} and {\it Feat} are non-empty,
finite or denumerably infinite sets.  The intuition is that these sets
denote the syntactic categories, the atomic values, and the features
that the linguist has chosen for some language.  For instance, $Cat$
could be $\{\mbox{S}, \mbox{NP}, \mbox{VP}, \mbox{V} \}$, $Atom$ might be
$\{sg, pl,
third, fem, masc \}$ and $Feat$ might be $\{subj, obj, pred, nb, case,
gd \}$.

Firstly we define $\cal{T}$.  As this is our mathematical embodiment of
c-structure (that is, a category labeled tree) we take it to be a pair
$\langle T, V_t \rangle$, where $T$ is a finite ordered tree and $V_t$
is a function from the set of tree nodes to $Cat$.  We will freely use
the usual tree terminology such as mother-of, daughter-of, dominates,
and so on.

Second, we take $\cal{F}$ to be a tuple of the form $\langle W,
\{f_{\alpha}\}_{\alpha \in Feat}, \mbox{{\it initial}}, \mbox{{\it
Final}}, V_{f} \rangle$, where $W$ is a finite, non-empty set of nodes;
$f_{\alpha}$ is a partial function from $W$ to $W$, for all $\alpha\in
\mbox{{\it Feat}}$; {\it initial} is a unique node in W such that any other
node $w$' of $W$ can be reached by applying a finite number of $f_{\alpha}$
to $initial$; {\it Final} is a non-empty set of nodes such that for all
$f_{\alpha}$ and all $w\in\mbox{{\it Final}}$, $f_{\alpha} (w)$ is
undefined; and $V_{f}$ is a function from {\it Final} to $\mbox{{\it
Atom}}$.  This is a standard way of viewing feature structures, and is
appropriate for LFG.

Finally, we take {\it zoomin}, the link between c-structure and f-structure
information, to be a partial function from $T$ to $W$.  This completes our
mathematical picture of LFG ontology.  It is certainly a precise picture
(all three components, and how they are related are well defined), but,
just as importantly, it is also a faithful picture;   models capture
the LFG ontology perspicuously.

\section{A Specification Language}

Although models pin down the essence of the LFG universe, our work has
only just begun.  For a start, not all models are created equal.
Which of them correspond to grammatical utterances of English? Of
Dutch?  Moreover, there is a practical issue to be addressed: how
should we go about saying which models we deem `good'? To put in
another way, in what medium should we specify grammars?

Now, it is certainly possible to talk about models using natural language
(as readers of this paper will already be aware) and for many purposes
(such as discussion with other linguists) natural language is undoubtedly
the best medium.  However, if our goal is to specify large scale grammars
in a clear, unambiguous manner, and to do so in such a way that our
grammatical analyses are machine verifiable, then the use of formal
specification languages has obvious advantages.  But which formal
specification language? There is no  single best answer: it depends on
one's goals.  However there are some important rules of thumb: one should
carefully consider the expressive capabilities required; and a judicious
commitment to simplicity and elegance will probably pay off in the long
run.  Bearing this advice in mind, let us consider the nature of LFG
grammars.

Firstly, LFG grammars impose constraints on  ${\cal T}$.
Context free rules are typically used for this purpose --- which means, in
effect, that constraints are being imposed on the `daughter of' and `sister
of' relations of the tree.  Secondly, LFG grammars impose general
constraints on various features in ${\cal F}$.  Such constraints (for
example the completeness constraint) are usually expressed in English and
make reference to specific features (notably {\it pred\/}).
Thirdly, LFG grammars impose constraints on {\it zoomin}.  As we have
already seen, this is done by annotating the context free rules with
equations.  These constraints regulate the interaction of the `mother of'
relation on ${\cal T}$, {\em zoomin}, and specific features in ${\cal F}$.

Thus a specification language adequate for LFG must be capable of talking
about the usual tree relations, the various features, and {\it zoomin}; it
must also be powerful enough to permit the statement of generalisations;
and it must have some mechanism for regulating the interaction between
${\cal T}$ and ${\cal F}$.  These desiderata can be met
by making use of a propositional language augmented with (1) modal
operators for talking about trees (2) modal operators for talking about
feature structures, and (3) a modal operator for talking about {\it
zoomin\/}, together with  a path equality construct for synchronising the
information flow between the two domains.  Let us build such a language.

Our language is called \logic\ and its primitive
symbols (with respect to a given
signature $\langle \mbox{{\it Cat}},
\mbox{{\it Atom}}, \mbox{{\it Feat\/}}
\rangle$) consists of (1) all items in {\it Cat} and {\it Atom}
(2) two constants, {\it
c-struct} and {\it f-struct}, (3) the Boolean connectives ({\it true}, {\it
false}, $\neg$, $\wedge$, $\rightarrow$, etc.), (4) three tree modalities
$\ua$, $\da$ and $\bullet$, (5) a modality $\langle \alpha \rangle$,
for each feature $\alpha\in \mbox{{\it Feat}}$,
(6) a synchronisation modality $\ftr{\zi}$,
(7) a path equality constructor $\peq$, together with (8)
the brackets $)$ and $($.

The basic well formed formulas (basic wffs) of \logic\ are:
$ \{\mbox{{\it true, false, c-struct, f-struct\/}}\}
\cup\mbox{{\it Cat\/}}\cup\mbox{{\it Atom\/}}\cup \mbox{{\it
Patheq\/}}$, where {\it Patheq\/} is defined as follows.  Let $t$, $t'$ be
finite (possibly null) sequences of the modalities $\ua$ and
$\da$, and let $f$, $f'$ be finite (possibly null) sequences of
feature modalities.
Then $t\ftr{\zi} f \peq t'\ftr{\zi}f'$ is in {\it Patheq\/},
and nothing else is.

The wffs of \logic\ are defined as follows: (1) all basic wffs are wffs,
(2) all Boolean combinations of wffs are wffs, (3) if $\phi$ is a wff
then so is $M\phi$, where $M\in \{ \langle{\alpha}\rangle: \alpha\in
\mbox{{\it Feat\/}} \}
\cup\{\ua, \da, \ftr{\zi}\}$
and (4) if $n>0$, and $\phi_1, \ldots, \phi_n$ are wffs, then so is
$\bullet (\phi_1,\ldots, \phi_n)$.  Nothing else is a wff.

Now for the satisfaction definition. We inductively
define a three place relation $\models$ which holds between
models ${\bf M}$, nodes $n$ and wffs $\phi$. Intuitively,
$\mo, n \models \phi$ means that the constraint $\phi$
holds at (is {\em true} at, is {\em satisfied} at)
the node $n$ in model ${\bf M}$.
The required inductive definition is as follows:

$$
\begin{array}{ll}

\mo, n \models \mbox{{\it true}} & \mbox{{\it always}}   \\

\mo, n \models \mbox{{\it false}} & \mbox{{\it never}}   \\

\mo, n \models \mbox{{\it c-struct}} & \ifff \\
\ \ \ \ \ n \; \txt{is a tree node} & \\

\mo, n \models \mbox{{\it f-struct}} & \ifff \\
\ \ \ \ \ n  \; \txt{is a feature structure node} &  \\

\mo, n \models c        & \ifff \\
\ \ \ \ \ V_{t}(n) = c, \; (\mbox{{\it for all}} \;
\ c  \in \mbox{{\it Cat}}) & \\

\mo, n \models a & \ifff \\
\ \ \ \ \ V_{f}(n) = a, \; (\mbox{{\it for all}} \;
a \in \mbox{{\it Atom}}) & \\

\mo, n \models \neg \phi & \ifff \\
\ \ \ \ \ not \; \ \mo, n \models \phi
& \\

\mo, n \models \phi \wedge \psi & \ifff \\
\ \ \ \ \  \mo, n \models
\phi\; \ and\; \ \mo, n \models \psi  & \\

\mo, n \models \ftr{\alpha} \phi & \ifff \\
\ \ \ \ \  f_{\alpha}(n) \;
exists\; and \;
\mo, f_{\alpha}(n) \models \phi & \\
\ \ \ \ \ (\txt{for all } \alpha \in Feat) &
\end{array}
$$

$$
\begin{array}{ll}

\mo, n \models \da \phi & \ifff \\
\ \ \ \ \ n \; \txt{ is a tree node with}  & \\
\ \ \ \ \ at\;   least\;  one\;  daughter\;  n' \;  such\;  that\; & \\
\ \ \ \ \ \mo, n'\models \phi &  \\

\mo, n \models \ua\phi & \ifff \\
\ \ \ \ \ n \; \txt{ is a tree node with} & \\
\ \ \ \ \ \txt{a mother node } m\; and\; & \\
\ \ \ \ \ \mo, m\models \phi & \\

\mo, n \models \ftr{zoomin} \phi
 & \ifff  \\
\ \ \ \ \ \txt{n is a tree node}, & \\
\ \ \ \ \  \zi(n) \; \txt{is defined}, and & \\
\ \ \ \ \ \mo, \zi(n) \models \phi & \\

\mo, n \models \bullet(\phi_1, \dots ,\phi_k) & \ifff \\
\ \ \ \ \ n \; \txt{ is a tree node with } & \\
\ \ \ \ \ exactly  \;  k\;  daughters\;  n_{1} \dots n_k\;  and\; & \\
\ \ \ \ \ \mo, n_1 \models \phi_1, \dots, $\mo$, n_k
\models \phi_k & \\

\mo, n \models
t\ftr{zoomin} f \peq\
t'  \ftr{zoomin} f'
 & \ifff  \\
\ \ \ \ \ n \; \txt{ is a tree node, and there is a} & \\
\ \ \ \ \  \txt{feature structure node} \; w \; \txt{such that}& \\
\ \ \ \ \ n ( S_t; zoomin; S_f ) w  \; and & \\
\ \ \ \ \ n ( S_{t'}; zoomin; S_{f'}) w  &
\end{array}
$$

For the most part the import of these clauses should be clear.  The
constants {\it true\/} and {\it false\/} play their usual role, {\it
c-struct\/} and {\it f-struct\/} give us `labels' for our two domains,
while the elements of {\it Cat\/} and {\it Atom} enable us to talk about
syntactic categories and atomic f-structure information respectively.  The
clauses for $\neg$ and $\wedge$ are the usual definitions of classical
logic, thus we have all propositional calculus at our disposal; as we shall
see, this gives us the flexibility required to formulate non-trivial
general constraints.  More interesting are the clauses for the modalities.
The unary modalities $\langle\alpha\rangle$, $\ua$, $\da$, and $\ftr{zoomin}$
and
the variable arity modality $\bullet$ give us access to the binary
relations important in formulating LFG grammars.  Incidentally, $\bullet$
is essentially a piece of syntactic sugar; it could be replaced by a
collection of unary modalities (see Blackburn and Meyer-Viol (1994)).
However, as the $\bullet$ operator is quite a convenient
piece of syntax for capturing the
effect of phrase structure rules, we have included it as a
primitive in \logic.

In fact, the only clause in the satisfaction definition which is at all
complex is that for $\peq$.  It can be glossed as follows.  Let $S_t$
and $S_{t'}$ be the path sequences through the tree corresponding to $t$
and $t'$ respectively, and let $S_f$ and $S_{f'}$ be the path sequences
through the feature structure corresponding to $f$ and $f'$ respectively.
Then $t \ftr{zoomin} f \peq\ t' \ftr{zoomin}f'$ is satisfied at a tree
node $t$ iff there is a feature structure node $w$ that can be reached from
$t$ by making both the transition sequence $S_t ;zoomin; S_f$ and the
transition sequence $S_{t'} ;zoomin; S_{f'}$.  Clearly, this construct is
closely related to the Kasper Rounds path equality (see Kasper and Rounds
(1990)); the principle difference is that whereas the Kasper Rounds
enforces path equalities within the domain of feature structures, the LFG
path equality enforces equalities between the tree domain and the feature
structure domain.

If ${\bf M}, n\models\phi$ then we say that $\phi$ is {\em satisfied}
in ${\bf M}$ at $n$.  If ${\bf M}, n\models\phi$ for all nodes $n$ in
\mo\ then we say that $\phi$ is {\em valid} in ${\bf M}$ and write
${\bf M}\models\phi$.  Intuitively, to say that $\phi$ is valid in ${\bf
M}$ is to say that the constraint $\phi$ holds universally; it is a
completely general fact about ${\bf M}$.  As we shall see in the next
section, the notion of validity has an important role to play in grammar
specification.

\section{Specifying Grammars}

We will now illustrate how \logic\ can be used to specify
grammars. The basic idea is as follows.  We write down a wff $\phi^G$
which expresses all our desired grammatical constraints.  That is, we
state in \logic\ which trees and feature structures are admissible,
and how tree and feature based information is to be synchronised;
examples will be given shortly.  Now, a model is simply a mathematical
embodiment of LFG sentence structure, thus those models ${\bf M}$ in
which $\phi^G$ is {\em valid\/} are precisely the sentence structures which
embody all our grammatical principles.

Now for some examples. Let's first consider how to write
specifications which capture the effect of schemata annotated grammar
rules.  Suppose we want to capture the meaning of rule (1) of Figure
1, repeated here for convenience:

$$\begin{array}{lccc}
\mbox{S} &      \longrightarrow         & \mbox{NP}     & \mbox{VP} \\
  &                             & (\uparrow \mbox{{\sc subj}}) = \downarrow
                                        & \uparrow = \downarrow \\
&&&\\
\end{array}$$
Recall that this annotated rule licenses  structures consisting of a
binary tree whose mother node $m$ is labeled S and whose daughter
nodes $n_1$ and $n_2$ are labeled NP and VP respectively; and where,
furthermore, the S and VP nodes (that is, $m$ and $n_2$) are related
to the same f-structure node $w$; while the NP node (that is, $n_1$)
is related to the node $w'$ in the f-structure that is reached by
making a \attr{subj} transition from $w$.

This is precisely the kind of structural constraint that \logic\ is
designed to specify.  We do so as follows:

$$\begin{array}{ll}
\mbox{S} \rightarrow &  \bullet(\mbox{NP}
\wedge \upa \ftr{\zi} \ftr{subj} \peq \ftr{\zi}, \\
  & \ \ \ \mbox{VP} \wedge \upa \ftr{\zi}\peq \ftr{\zi} )
\end{array}$$
\noindent
This formula
is satisfied in a model \mo\ at any node $m$ iff $m$ is labeled with the
category $S$, has exactly two daughters $n_1$ and $n_2$ labeled with
category NP and VP respectively.  Moreover, $n_1$ must be associated
with an f-structure node $w'$ which can also be reached by making a
$\ftr{subj}$ transition from the f-structure node $w$ associated with
the mother node of $m$.  (In other words, that part of the f-structure
that is associated with the NP node is re-entrant with the value of the
subj feature in the f-structure associated with the S node.) And
finally, $n_2$ must be associated with that f-structure node $w$ which
$m$.  (In other words, the part of the f-structure that is associated
with the VP node is re-entrant with that part of the f-structure which is
associated with the S node.)

In short, we have captured the effect of an annotated rule purely
declaratively. There is no appeal to any construction algorithm;
we have simply stated how we want the different pieces to fit
together.  Note that $\bullet$ specifies local tree admissibility
(thus obviating the need for rewrite rules), and $\ftr{\zi}$, $\ua$
and $\peq$ work together to capture the effect of $\downarrow$ and
$\uparrow$.

In any realistic LFG grammar there will be several --- often many --- such
annotated rules, and acceptable c-structures are those
in which each non-terminal node is licensed
by one of them.  We specify this as follows.  For each such rule we
form the analogous \logic\ wff $\phi^{r}$ (just as we did in the previous
example) and then we form the {\em disjunction} $\bigvee\phi^{r}$ of all
such wffs.  Now, any non-terminal node
in the c-structure should satisfy one of these disjunctions (that is, each
sub-tree of {\it c-struct} must be licensed by one of these conditions);
moreover
the disjunction is irrelevant to the terminal nodes of {\it
c-struct\/} and all the nodes in {\it f-struct\/}. Thus we
demand that the following conditional statement be valid:
$$(\mbox{{\it c-struct}}\wedge \da\mbox{{\it true\/}})\rightarrow
\bigvee\phi^{r}.$$
This says that {\it if\/} we are at a {\it c-struct\/} node which has at
least one daughter (that is, a non-terminal node) {\it then\/} one of the
subtree licensing disjuncts (or `rules') must be satisfied there.  This
picks precisely those models in which all the tree nodes are appropriately
licensed.  Note that the statement is indeed {\em valid} in such models: it
is true at all the non-terminal nodes, and
is vacuously satisfied at terminal tree nodes and nodes of {\it
f-struct\/}.

We now turn to the second main component of LFG, the well formedness
conditions on f-structures.

Consider first the uniqueness principle.  In essence, this principle
states that in a given f-structure, a particular attribute may have at
most one value.  In \logic\ this restriction is `built in': it follows
from the choices made concerning the mathematical objects composing
models.  Essentially, the uniqueness principle is enforced by two
choices.  First, $V_{f}$ associates atoms only with final nodes of
f-structures; and as $V_f$ is a function, the atom so
associated is unique.  In effect, this hard-wires prohibitions against
constant-compound and constant-constant clashes into the semantics of
\logic.  Second, we have modeled features as partial functions on the
f-structure nodes -- this ensures that any complex valued attribute is
either undefined, or is associated with a unique sub-part of the
current f-structure.  In short, as required, any
attribute will have at most one value.

We turn to the completeness principle. In LFG, this
applies to
a (small) finite number of attributes (that is,
transitions in the feature
structure). This collection includes the
grammatical functions (e.g. {\it subj\/},
{\it obj\/}, {\it iobj\/}) together with some
longer transitions such as $obl;obj$
and $to;obj$.
Let {\it GF} be a metavariable
over the modalities corresponding to the
elements of this set,
thus {\it GF} contains such items as
$\langle \mbox{{\it subj\/}}\rangle$,
$\langle \mbox{{\it obj\/}}\rangle$,
$\langle \mbox{{\it iobj\/}}\rangle$,
$\langle \mbox{{\it obl\/}}\rangle\langle \mbox{{\it obj\/}}\rangle$ and
$\langle \mbox{{\it to\/}}\rangle\langle \mbox{{\it obj\/}}\rangle$.
Now, the completeness principle requires that
any of these features
appearing as an attribute in the value of the
\attr{pred} attribute must also appear as an attribute of the f-structure
immediately containing this \attr{pred} attribute, and this
recursively.  The following wff is valid on precisely those models
satisfying the completeness principle:

\[
\ftr{pred} \mbox{{\it GF} } true \rightarrow \mbox{{\it GF} } true.
\]

Finally, consider the counterpart of the completeness principle, the
coherence principle.  This applies to the same attributes as the
completeness principle and requires that whenever they occur in an
f-structure they must also occur in the f-structure associated with its
\attr{pred} attribute.  This is tantamount to demanding the validity of the
following formula:

\[
(\mbox{{\it GF} } true \wedge
\ftr{pred}true)\rightarrow \ftr{pred} \mbox{{\it GF} } true
\]

\section{Conclusion}

The discussion so far should have given the reader some idea of how to
specify LFG grammars using ${\cal L}$.  To conclude we would like to
discuss $=_c$ definitions.  This topic bears on an important general
issue: how are the `dynamic' (or `generative', or `procedural')
aspects of grammar to be reconciled with the `static', (or
`declarative') model theoretic world view.

The point is this. Although the LFG equations discussed so far were
{\it defining equations}, LFG also allows  so-called {\it
constraining equations}  (written $=_c$). Kaplan and Bresnan explain the
difference as follows.  Defining equations allow  a feature-value
pair to be inserted into an f-structure providing no conflicting
information is present. That is, they add a feature value pair to any
consistent f-structure. In contrast, constraining equations are
intended to constrain the value of an already existing feature-value
pair. The essential difference is that constraining equations require
that the feature under consideration already has a value, whereas
defining equations apply independently of the feature value
instantiation level.

In short, constraining equations are essentially a global check on
completed structures which require the presence of certain feature
values.  They have an eminently procedural character, and there is no
obvious way to handle this idea in the present set up.  The bulk of
LFG involves stating constraints about a single model, and \logic\
is well equipped for this task, but constraining equations involve
looking at the structure of other possible parse trees.  (In this
respect they are reminiscent of the feature specification defaults of
GPSG.) The approach of the present paper has been driven by the view
that (a) models capture the essence of LFG ontology, and, (b) the task
of the linguist is to explain, {\em in terms of the relations that
exist within a single model}, what grammatical structure is.  Most of
the discussion in Kaplan and Bresnan (1982) is conducted in such
terms.  However constraining equations broaden the scope of the
permitted discourse; basically, they allow implicit appeal to possible
derivational structure.  In short, in common with most of the
grammatical formalisms with which we are familiar, LFG seems to have a
{\em dynamic} residue that resists a purely declarative analysis.
What should be done?

We see three possible responses.  Firstly, we note that the model
theoretic approach can almost certainly be extended to cover
constraining equations.  The move involved is analogous to the way
first order logic (a so-called `extensional' logic) can be extended to
cope with intensional notions such as belief and necessity.  The basic
idea --- it's the key idea underlying first order Kripke semantics ---
is to move from dealing with a single model to dealing with a
collection of models linked by an accessibility relation.  Just as
quantification over possible states of affairs yields analyses of
intensional phenomena, so quantification over related models could
provide a `denotational semantics' for $=_c$.  Preliminary work
suggests that the required structures have formal similarities to the
structures used in preferential semantics for default and
non-monotonic reasoning.  This first response seems to be a very
promising line of work: the requisite tools are there, and the
approach would tackle a full blooded version of LFG head on.  The
drawback is the complexity it introduces into an (up till now) quite
simple story.  Is such additional complexity really needed?

A second response is to admit that there is a dynamic residue, but to deal
with it in overtly computational terms.  In particular, it may be
possible to augment our approach with an explicit operational semantics,
perhaps the evolving algebra approach adopted by Moss and Johnson (1994).
Their approach is attractive, because it permits a computational treatment
of dynamism that abstracts from low level algorithmic details.  In short,
the second strategy is a `divide and conquer' strategy: treat structural
issues using model theoretic tools, and procedural issues with
(revealing) computational tools.  It's worth remarking that this second
response is not incompatible with the first; it is common to provide
programming languages with both a denotational and an operational
semantics.

The third strategy is both simpler and more speculative.  While it
certainly seems to be the case that LFG (and other `declarative'
formalisms) have procedural residues, it is far from clear that these
residues are necessary.  One of the most striking features of LFG (and
indeed, GPSG) is the way that purely structural (that is, model theoretic)
argumentation dominates.  Perhaps the procedural aspects are there more or
less by accident? After all, both LFG and GPSG drew on (and developed) a
heterogeneous collection of traditional grammar specification tools, such
as context free rules, equations, and features.  It could be the case such
procedural residues as $=_c$ are simply an artifact of using the wrong
tools for talking about models.  If this is the case, it might be highly
misguided to attempt to capture $=_c$ using a logical specification
language.  Better, perhaps, would be to draw on what is good in LFG and to
explore the logical options that arise naturally when the model theoretic
view is taken as primary.  Needless to say, the most important task that
faces this third response is to get on with the business of writing
grammars; that, and nothing else, is the acid test.

It is perhaps worth adding that at present the authors simply do not know
what the best response is.  If nothing else, the present work has made very
clear to us that the interplay of static and dynamic ideas in generative
grammar is a delicate and complex matter which only further work can resolve.

\end{document}